\documentclass{article}
\usepackage{amsmath}
\usepackage{multirow}
\usepackage{amssymb}
\usepackage{graphicx}
\usepackage{subfigure}
\usepackage{color}
\usepackage[preprint]{spconfa4}

\copyrightnotice{978-1-6654-3864-3/21/\$31.00 ©2021 IEEE}

\let\OLDthebibliography\thebibliography
\renewcommand\thebibliography[1]{
  \OLDthebibliography{#1}
  \setlength{\parskip}{0pt}
  \setlength{\itemsep}{0pt plus 0.3ex}
}

\begin{document}\sloppy

% Example definitions.
% --------------------
\def\x{{\mathbf x}}
\def\L{{\cal L}}

% Title.
% ------
\title{Confidence-guided Adaptive Gate and Dual Differential Enhancement For Video Salient Object Detection}
%
% Single address.
% ---------------
% \name{Anonymous ICME submission}
%Address and e-mail should NOT be added in the submission paper. They should be present only in the camera ready paper. 
% \address{}

\name{Peijia Chen$^1$, Jianhuang Lai$^{1,2,3\ast}$\thanks{$^{\ast}$Corresponding Author. This project is supported by the Natural Science Foundation of China (62076258).}, Guangcong Wang$^1$, Huajun Zhou$^1$}
\address{$^1$School of Computer Science and Engineering, Sun Yat-sen University, China \\ $^2$Guangdong Province Key Laboratory of Information Security Technology, China \\ $^3$Key Laboratory of Machine Intelligence and Advanced Computing, Ministry of Education, China \\ chenpj8@mail2.sysu.edu.cn, stsljh@mail.sysu.edu.cn}

\maketitle

%Recently, temporal information largely improves the performance of video salient object detection by introducing a temporal branch to provide explicit motion cues from optical flow images.
%temporal coherence

%However, the misleading information both from the spatial and temporal branches would interfere with the distinguishment of the salient objects. 

%suppress the misleading information from the spatial and temporal branches
%: (i) fast motion, (ii) illumination changes, (iii) occlusions, and (iv) untextured regions
%motion discontinuities, large displacements
% low-contrast attentive objects
% (e.g., RGB images)
% However, spatial and temporal cues are often unreliable in real-world scenarios, such as low-contrast foreground, fast motion, large displacements, motion discontinuities, and illumination changes.
\begin{abstract}
Video salient object detection (VSOD) aims to locate and segment the most attractive object by exploiting both spatial cues and temporal cues hidden in video sequences. However, spatial and temporal cues are often unreliable in real-world scenarios, such as low-contrast foreground, fast motion, and multiple moving objects. To address these problems, we propose a new framework to adaptively capture available information from spatial and temporal cues, which contains Confidence-guided Adaptive Gate (CAG) modules and Dual Differential Enhancement (DDE) modules. For both RGB features and optical flow features, CAG estimates confidence scores supervised by the IoU between predictions and the ground truths to re-calibrate the information with a gate mechanism. DDE captures the differential feature representation to enrich the spatial and temporal information and generate the fused features. Experimental results on four widely used datasets demonstrate the effectiveness of the proposed method against thirteen state-of-the-art methods.
\end{abstract}
% With CAG and DDE, our proposed model filters unavailable cues and enhance the xxx.

% optical flow:
% https://link.springer.com/chapter/10.1007/978-3-030-32583-1_12
% http://bigwww.epfl.ch/publications/fortun1501.pdf

\section{Introduction}
% background
Video salient object detection (VSOD) aims to locate and segment the most attractive object in video sequences, which is widely used as an important pre-processing step to reduce computational burdens for many high-level computer vision tasks, such as video compression \cite{itti2004automatic}, video captioning \cite{pan2017video} and person re-identification \cite{zhao2013unsupervised}.

\begin{figure}[!t] %加*的作用是跨栏（双栏和单栏latex的区别）
% \flushleft
% %每列占整个文档的文本宽度的0.2，每列两张图像，两列图像，这些可以自行设置
% \subfigure[RGB]{
% \begin{minipage}[b]{0.075\textwidth}
% \includegraphics[width=0.65in,height=0.4in]{simple/5_1.png} \\ 
% \includegraphics[width=0.65in,height=0.4in]{simple/3_1.jpg} \\
% \includegraphics[width=0.65in,height=0.4in]{simple/4_1.jpg}
% \end{minipage}
% }
% \hspace{0.02cm}
% \subfigure[OF]{
% \begin{minipage}[b]{0.075\textwidth}
% \includegraphics[width=0.65in,height=0.4in]{simple/5_2.png} \\ 
% \includegraphics[width=0.65in,height=0.4in]{simple/3_2.jpg} \\
% \includegraphics[width=0.65in,height=0.4in]{simple/4_2.jpg}
% \end{minipage}
% }
% \hspace{0.02cm}
% \subfigure[MGA]{
% \begin{minipage}[b]{0.075\textwidth}
% \includegraphics[width=0.65in,height=0.4in]{simple/5_4.png} \\ 
% \includegraphics[width=0.65in,height=0.4in]{simple/3_4.png} \\
% \includegraphics[width=0.65in,height=0.4in]{simple/4_4.png}
% \end{minipage}
% }
% \hspace{0.02cm}
% \subfigure[CINet]{
% \begin{minipage}[b]{0.075\textwidth}
% \includegraphics[width=0.65in,height=0.4in]{simple/5_5.png} \\ 
% \includegraphics[width=0.65in,height=0.4in]{simple/3_5.png} \\
% \includegraphics[width=0.65in,height=0.4in]{simple/4_5.png}
% \end{minipage}
% }
% \hspace{0.02cm}
% \subfigure[GT]{
% \begin{minipage}[b]{0.075\textwidth}
% \includegraphics[width=0.65in,height=0.4in]{simple/5_3.png} \\ 
% \includegraphics[width=0.65in,height=0.4in]{simple/3_3.png} \\
% \includegraphics[width=0.65in,height=0.4in]{simple/4_3.png}
% \end{minipage}
% }
\includegraphics[width=0.48\textwidth]{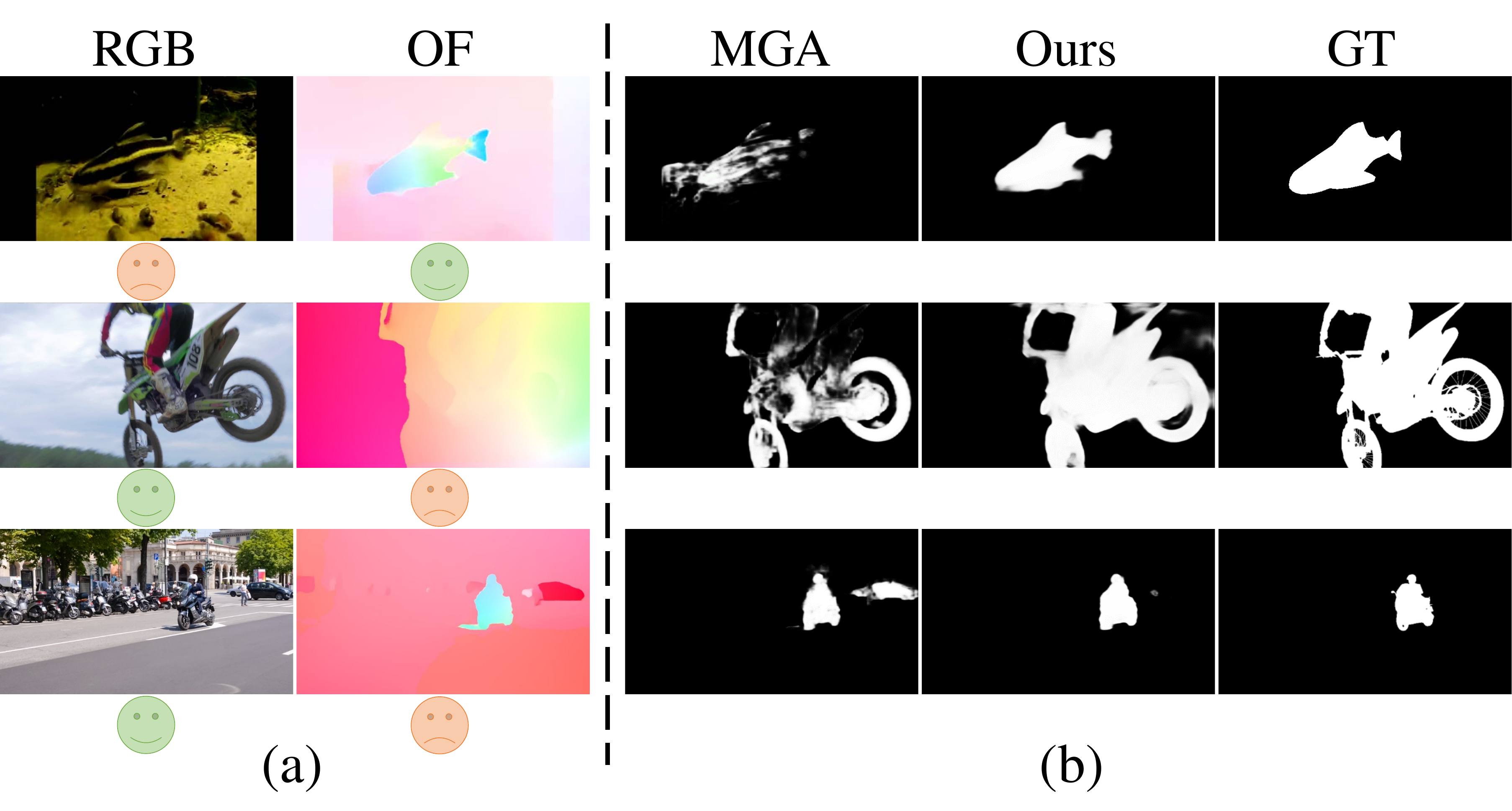}	
\caption{The challenges for video salient object detection in scenes of low-contrast foreground (the first row), fast motion (the second row) and multiple moving objects (the third row). OF denotes the optical flow image and GT denotes the ground truth.}
\label{fig:simple}
\end{figure}

% related work
Different from salient object detection (SOD) that focuses on predicting saliency maps by exploiting spatial information extracted from one single image, VSOD further exploits temporal information hidden in video sequences. Although VSOD exploits more information to predict saliency maps, it also brings extra challenges, as shown in Figure \ref{fig:simple}. \textbf{First}, spatial cues hidden in every single frame are often hard to be exploited when foreground and background share a similar feature representation. The first row shows that the RGB images of low contrast between the salient objects and backgrounds would bring in misleading information to predict background objects wrongly. \textbf{Second}, temporal cues hidden between different frames could be disturbed by fast motion, large displacements and illumination changes. The second row shows that the motocross is distinctive from the background in the RGB image but noisy in the optical flow image, which leads to the absence of the rider in the saliency map predicted by MGA. The third row shows that even the temporal information from the accurate optical flow images would confuse the spatial information in scenes of several moving objects. Driven by these two challenges, one would ask: how can we establish a model to automatically capture available spatial and temporal cues while suppress noisy ones?

% Based on the above observations, the spatial and temporal information share the same importance to predict the saliency maps, and the main challenge of VSOD is how to predict the saliency maps from the noisy spatial and temporal information.

% unreliable low-contrast attentive objects detected by spatial cues (e.g., RGB images) images and multiple ambiguous objects detected by temporal cues (e.g., optical flow images).

\begin{figure*}[!t]
\centering
\includegraphics[width=0.85\textwidth]{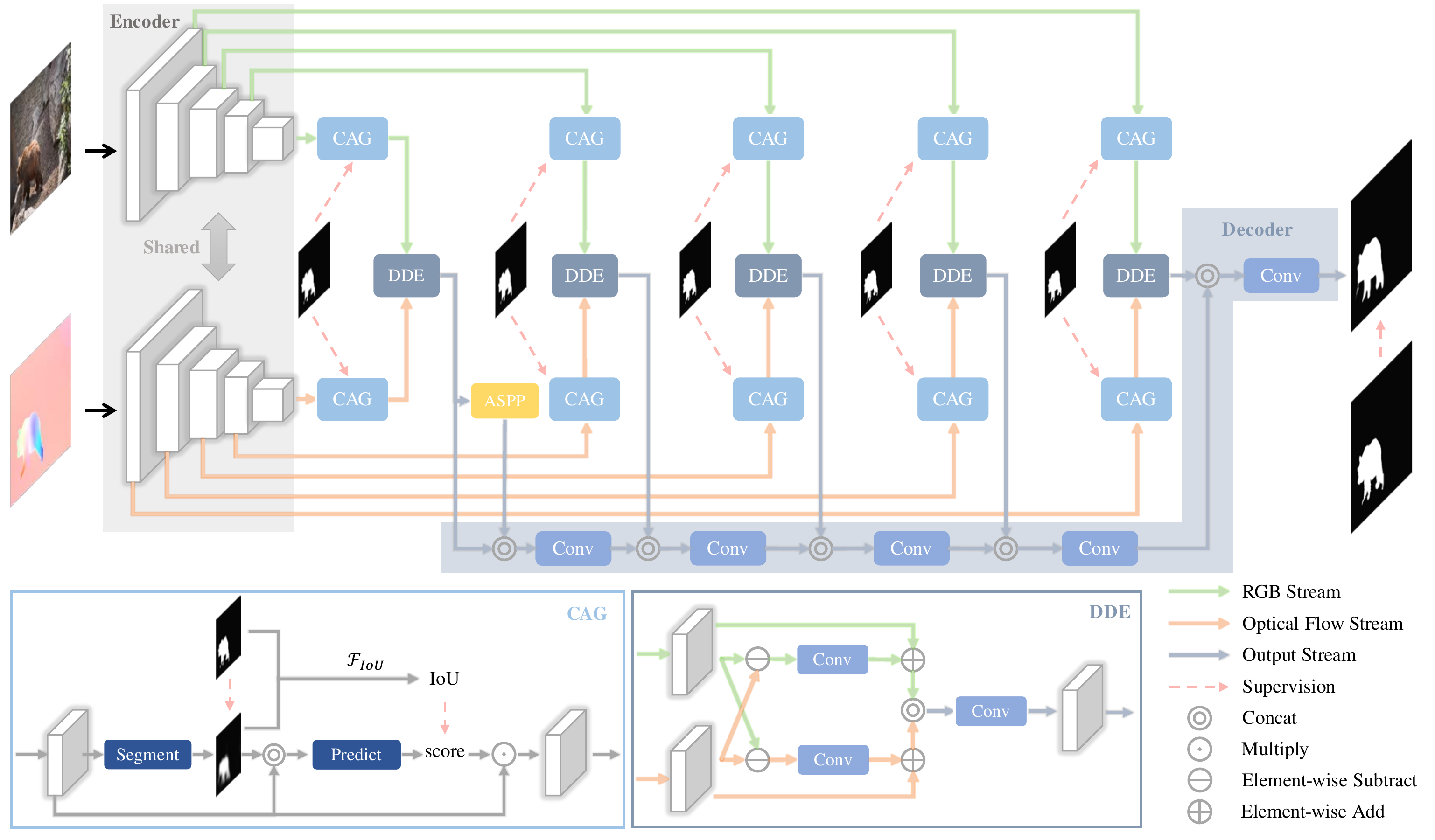}	
\caption{The overall architecture of the proposed model.}
% It consists of a shared encoder, an output decoder, Confidence-guided Adaptive Gate (CAG) modules and Dual Differential Enhancement (DDE) modules. The shared encoder extracts multi-level features from RGB and optical flow images. Next, these features are balanced by CAG and merged by DDE at different layers. Finally, five features at different layers are gradually combined by the output decoder to generate the final saliency map.
\label{fig:overview}
\end{figure*}
% It consists of a shared encoder, an output decoder, Confidence-guided Adaptive Gate (CAG) modules and Dual Differential Enhancement (DDE) modules. The shared encoder extract multi-level features from RGB and optical flow images. These features are balanced by CAG and merged by DDE at different layers. Finally, five features at different layers are gradually combined by the output decoder to generate the final saliency map.

The existing methods may offer a partial solution to this problem, which can be roughly classified into four groups, i.e., hand-crafted VSOD, long-term memory VSOD, attention VSOD, and parallel VSOD. \textbf{First}, hand-crafted VSOD methods try to combine the spatial information with motion cues based on the prior knowledge, such as the spatio-temporal background prior \cite{xi2016salient} and low-rank coherency \cite{chen2017video}, which yield poor performance limited by the hand-crafted low-level features. \textbf{Second}, long-term memory VSOD methods \cite{li2018flow,song2018pyramid,fan2019shifting} extract the spatial information from the single image separately and model the temporal information through convolutional memory units such as ConvLSTM. \textbf{Third}, attention VSOD methods use a non-local mechanism to capture temporal information across several images \cite{2020Pyramid}. The second and third groups are cascaded models that extract features for each frame and then model the dependence relationship between video sequences based on the extracted features. These ``first spatial and then temporal" cascaded models cannot adaptively capture available cues from both spatial and temporal cues to cooperatively predict saliency maps. \textbf{Fourth}, instead of using cascaded ways, parallel VSOD methods \cite{li2019motion} often adopt a two-stream framework where one stream is to extract features from every single frame and another stream is to process temporal information independently by using optical flow images. However, the existing parallel VSOD methods simply fuse spatial and temporal information without considering the confidence of the spatial and temporal information. 

% from optical flow images, which could deteriorate the accuracy of saliency maps. Besides, it only enriches the spatial information with temporal information, while these two information should interact with each other to achieve better performance due to their complementarity?. %salient

% to capture motion cues. 

% and develop a branch that process the temporal information independently to attend and enhancement the spatial branch.

% the optical flow to sense the motion cues and develop a branch that process the temporal information independently to attend and enhancement the spatial branch.

% our method, top-down, confidence-aware
% complement them in an interactive way.
% to reduce the interference by the misleading information,
% suppress the features with its confidence score.
To address the above issues, we propose a new framework that adaptively captures available spatial and temporal information to predict completely saliency maps. Specifically, we introduce a Confidence-guided Adaptive Gate (CAG) module that can estimate the confidence score of the input features by calculating IoU value. With the confidence score, CAG passes available information and suppresses the noises from spatial and temporal cues. We also propose a Dual Differential Enhancement (DDE) module that focuses on capturing differential information between spatial and temporal cues and generates the fused features.
% enhance the diff full interaction between the spatial and temporal information to obtain discriminative features for

Overall, our main contributions are summarized as follows:
\begin{itemize}
  \item We propose a new framework to accurately predict the saliency maps for video salient object detection, which adaptively captures available information from spatial and temporal cues.
  \item We propose a Confidence-guided Adaptive Gate (CAG) module to suppress low-confidence information and a Dual Differential Enhancement (DDE) module to exploit discriminative features enhanced by the differential information from the spatial and temporal dual branch.  %by taking advantage of the complementarity between the spatial and temporal information.
  \item Experiments on four widely used datasets demonstrate the superiority of our proposed method against state-of-the-art methods. 
\end{itemize}

% Extensive experiments are conducted to demonstrate the effectiveness of our method.
%, which could diminish the influence of noise from

\section{Proposed Method}
% overview
In this section, we propose a Confidence-guided Adaptive Gate and Dual Differential Enhancement (CAG-DDE) method for video salient object detection, which passes the available information while suppresses the unreliable information from spatial and temporal cues and enhances them with differential information to completely segment salient objects. As shown in Figure \ref{fig:overview}, the overall framework of the proposed method is built on the encoder-decoder architecture, which includes a shared encoder, an output decoder, a number of Confidence-guided Adaptive Gate (CAG) modules and Dual Differential Enhancement (DDE) modules.
% Given a pair of RGB and optical flow images (visualizations of optical flow), the proposed network completely segments salient objects by analyzing the spatial and temporal information from the RGB and optical flow branches comprehensively. For the shared encoder, we adopt ResNet-101 \cite{he2016deep} to extract different level features, denoted as $C_i$ (colorful) for RGB features and $O_i$ (optical) for optical flow features ($i \in (1, 2, 3, 4, 5)$ indexes different levels). 
Given a pair of RGB and optical flow images (visualization of optical flow), the shared encoder outputs the RGB features and optical flow features at five different layers. Let $E_i^{RGB}$ and $E_i^{OF}$ denote RGB features and optical flow features, respectively. Here, $i$ indicates the $i$-th layer, $1\leq i \leq 5$. Both $E_i^{RGB}$ and $E_i^{OF}$ are re-calibrated with their confidence scores estimated by the CAG modules, and then merged by the DDE modules to obtain the decoder features $D_i$, which can be formulated as
\begin{eqnarray}
D_i = \mathcal{F}_{DDE}(\mathcal{F}_{CAG}(E_i^{RGB}),\mathcal{F}_{CAG}(E_i^{OF})),
\end{eqnarray}
where $\mathcal{F}_{CAG}$ and $\mathcal{F}_{DDE}$ denote a CAG operator and a DDE operator. Next, five decoder features $\{D_i\}_{i=1}^5$ at different layers are gradually combined by the output decoder. Following \cite{li2019motion}, we integrate an atrous spatial pyramid pooling (ASPP) module after the DDE module at the fifth layer. The output decoder is formulated as
\begin{eqnarray}
% F^{HLLC}=\left\{  \begin{array}{rcl} F_L       &      & {0      <      S_L}\\ F^*_L     &      & {S_L \leq 0 < S_M}\\ F^*_R     &      & {S_M \leq 0 < S_R}\\ F_R       &      & {S_R \leq 0} \end{array} \right.
&&\hspace{-3.6em}D_i^{'} = \left\{
\begin{array}{rl}
\mathcal{F}_{conv}(D_i\circledcirc \mathcal{F}_{upsample}(D_{i+1}^{'})) & i = 1, 2, 3, 4
\\
\mathcal{F}_{conv}(D_5\circledcirc \mathcal{F}_{ASPP}(D_5)) & i = 5
\end{array}
\right.
\end{eqnarray}
where $\circledcirc$ is a concatenation operator. $\mathcal{F}_{conv}$ and $\mathcal{F}_{upsample}$ are the operators of convolution and upsampling. Finally, we upsample the final decoder features as the predicted saliency map, which can be formulated as
\begin{eqnarray}
P_f = \mathcal{F}_{upsample}(D_1^{' }).
\end{eqnarray}

% \begin{eqnarray}
% &&\mathcal{F}_{CAG}(E_i) = E_i \cdot s_i, \\
% &&s_i = \sigma(\mathcal{F}_{pred}(\mathcal{F}_{seg}(E_i)\circledcirc E_i))
% \end{eqnarray}

% \begin{eqnarray}
% &&\mathcal{F}_{DDE}(E_i, E_i) = \mathcal{F}_{conv}(\mathcal{F}_{c}(E_i, E_i) \circledcirc \mathcal{F}_{c}(E_i, E_i)), \\
% &&\mathcal{F}_{c}(X, Y) = \mathcal{F}_{conv}(X-Y)+Y
% \end{eqnarray}

% We first provide an overview of our proposed framework in Section \ref{subsec:overall}. In Section \ref{subsec:CAG}, we provide a detailed introduction about the Adaptive Gate Module. In Section \ref{subsec:DDE}, we propose a Feature Interaction Module to incorporate the spatial and temporal information interactively.
% \subsection{Overall network architecture}
% \label{subsec:overall}

\subsection{Confidence-guided Adaptive Gate Module}
\label{subsec:CAG}
% motivation
% Video salient object detection needs to locate the most attractive objects in video sequences depending on not only the spatial features but temporal features. However, both the spatial and temporal features mix with unreliable information inevitably.
% For instance, low-contrast RGB images confuse the salient object and background, and optical flow estimation may yield poor performance in complex movement scenes.
In real-world scenarios, both the spatial and temporal images contain unreliable information inevitably. How to measure the reliability of the informative features and noisy features is a key to the VSOD problem. To address the problem, we propose a Confidence-guided Adaptive Gate (CAG) module, which predicts the confidence score to represent the reliability of the features and re-calibrate the features, as shown in the bottom left of Figure \ref{fig:overview}.

Our CAG module is composed of two sub-networks, the segmentation sub-network and confidence score prediction sub-network. The segmentation sub-network consists of three convolution layers, which is used to predict the saliency map $P_i$ supervised by the ground truth $G_i$. The confidence score prediction sub-network consists of three convolution layers and one global average pooling layer, which aims to explicitly model the confidence score. We concatenate the predicted saliency map and the input features as the input of the confidence score prediction sub-network. Inspired by a segmentation quality metric Intersection over Union (IoU), we quantify the confidence score as the IoU between the saliency map $P_i$ and the ground truth $G_i$.
% formula
Given the input feature $E_i$, we can obtain the confidence score by
\begin{eqnarray}
s_i = \sigma(\mathcal{F}_{pred}(\mathcal{F}_{seg}(E_i)\circledcirc E_i)),
\end{eqnarray}
where $\sigma$ is a sigmoid function that scales the confidence score to (0,1). $\mathcal{F}_{seg}$ and $\mathcal{F}_{pred}$ refer to the segmentation sub-network and confidence score prediction sub-network, respectively.

Under the guidance of the predicted confidence scores, our method adaptively re-calibrates the features based on a gate mechanism, which is given by
\begin{eqnarray}
\mathcal{F}_{CAG}(E_i) = E_i \cdot s_i(E_i),
\end{eqnarray}
where $\cdot$ denotes the scalar multiplication. For each layer of the encoder, we employ a segmentation sub-network and a confidence score prediction sub-network to estimate the confidence scores of features at different levels.

\subsection{Dual Differential Enhancement Module}
\label{subsec:DDE}
% The CAG aims at balancing the contributions of RGB and optical flow features thereby suppress the misleading information, while the feature interaction module (DDE) are proposed to interact these two features with each other to gain comprehensive information. 

% Each of these two features focuses on only one cue leading to incomplete salients objects while these two features are strongly complementary to each other.

% Through the confidence-guided adaptive gate module, we obtain the re-calibrated RGB features and optical flow features.
Color saliency obtained by the RGB feature and motion saliency obtained by the optical flow feature are strongly complementary with each other. However, most complementary information is hidden in the difference between RGB and optical flow features. To make full use of their complementarity, we propose a Dual Differential Enhancement (DDE) module to discover differential information between the RGB and optical flow features, as shown at the bottom right of Figure \ref{fig:overview}. For each branch, we extract the differential information by subtracting the shared information and try to focus on learning branch-specific (spatial or motion) information. We then enhance the original information by adding the differential information. After the dual differential enhancement between the RGB and optical flow features, the merged feature is computed by
% \begin{small}
\begin{eqnarray}
&&\hspace{-3.5em}\mathcal{F}_{DDE} = \mathcal{F}_{conv}(\mathcal{F}_{e}(R_i^{RGB}, R_i^{OF}) \circledcirc \mathcal{F}_{e}(R_i^{OF}, R_i^{RGB})),\\
&&\hspace{1em}\mathcal{F}_{e}(X, Y) = \mathcal{F}_{conv}(Y-X)+X
\end{eqnarray}
% \end{small}
where $R_i^{RGB}$ and $R_i^{OF}$ denote the re-calibrated RGB and optical flow features, respectively. $\mathcal{F}_{DDE}$ refers to an operator of dual differential enhancement module and $\mathcal{F}_{e}$ refers to an operator of differential enhancement for a single branch.

\begin{figure*}[!t] %加*的作用是跨栏（双栏和单栏latex的区别）
\centering
%每列占整个文档的文本宽度的0.2，每列两张图像，两列图像，这些可以自行设置
\subfigure[Image]{
\begin{minipage}[b]{0.10\textwidth}
\includegraphics[width=0.82in,height=0.5in]{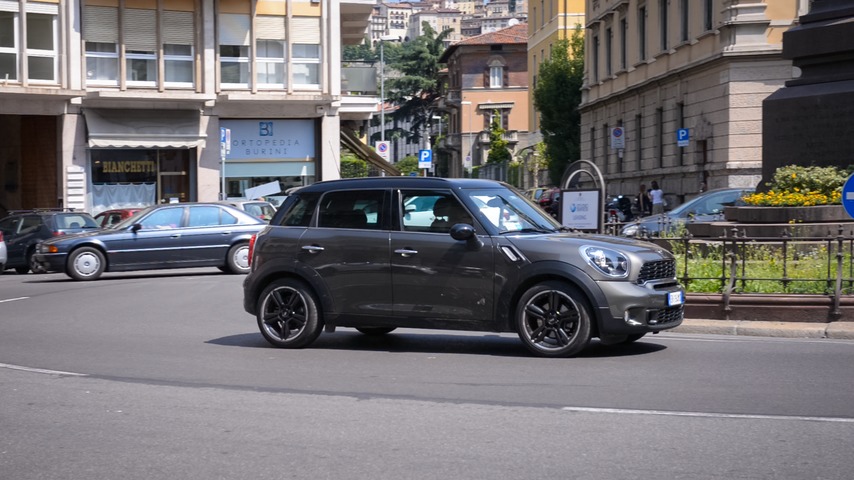} \\ 
\includegraphics[width=0.82in,height=0.5in]{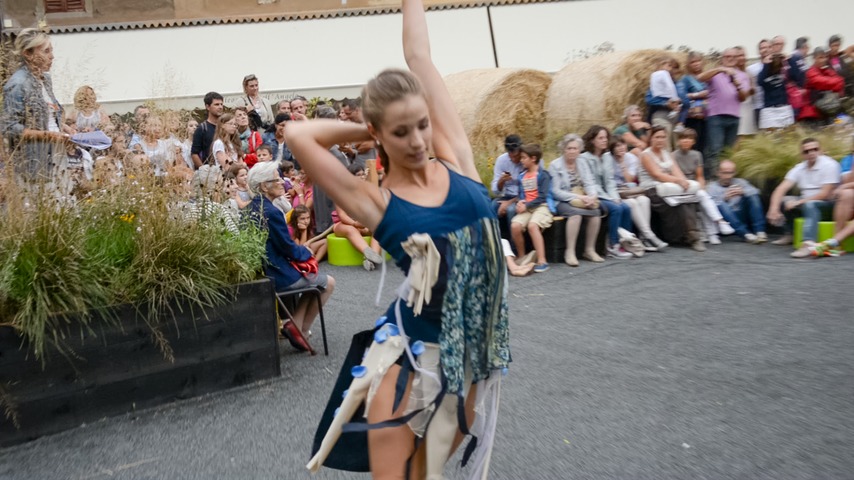} \\ 
\includegraphics[width=0.82in,height=0.5in]{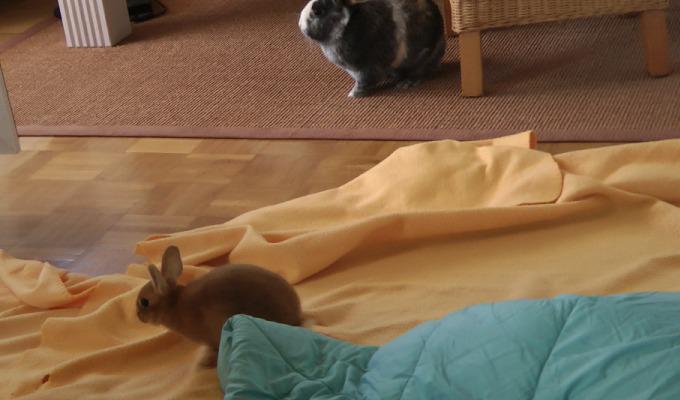} \\ 
\includegraphics[width=0.82in,height=0.5in]{img/3_1.png} \\ 
\includegraphics[width=0.82in,height=0.5in]{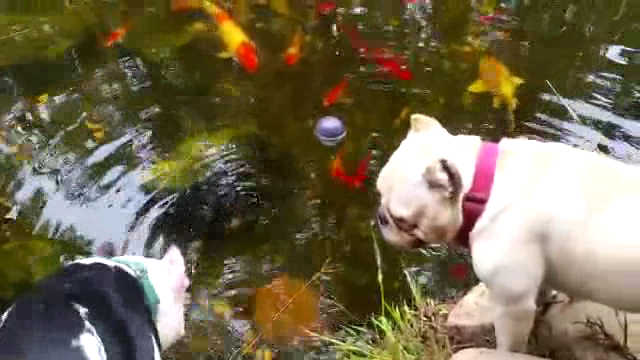}
\end{minipage}
}
\hspace{0.02cm}
\subfigure[GT]{
\begin{minipage}[b]{0.10\textwidth}
\includegraphics[width=0.82in,height=0.5in]{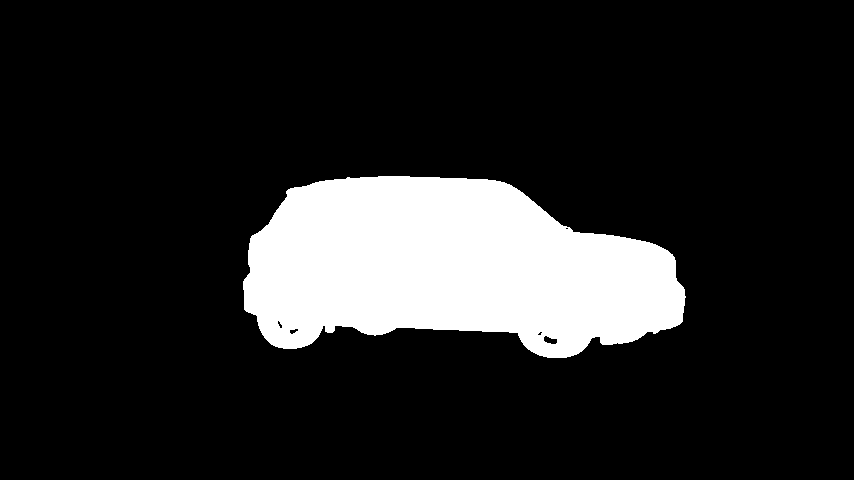} \\ 
\includegraphics[width=0.82in,height=0.5in]{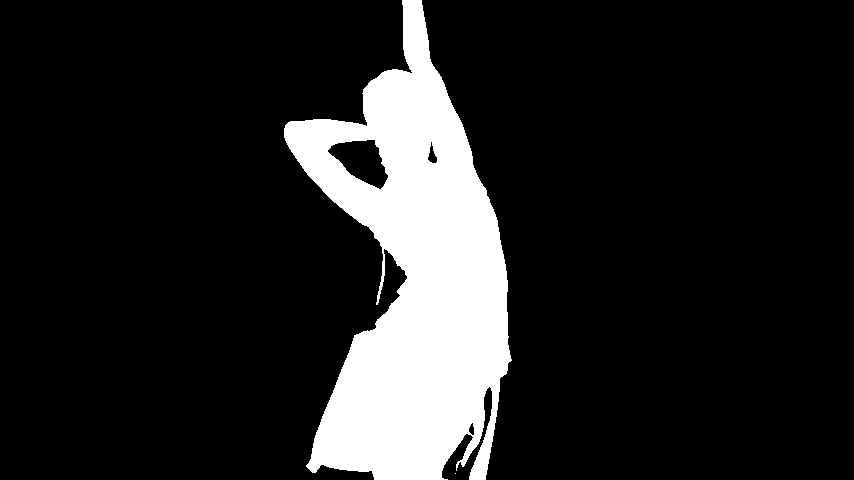} \\ 
\includegraphics[width=0.82in,height=0.5in]{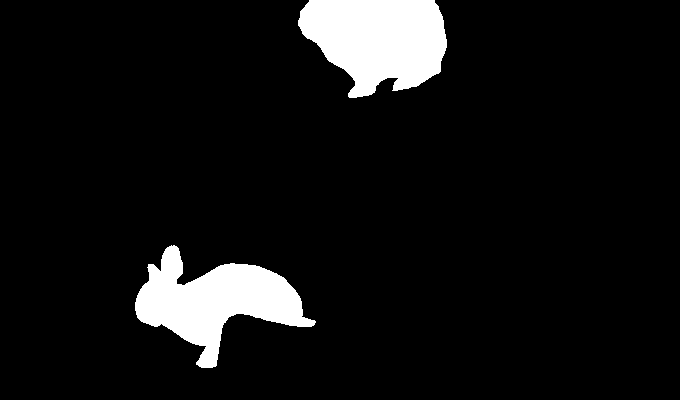} \\ 
\includegraphics[width=0.82in,height=0.5in]{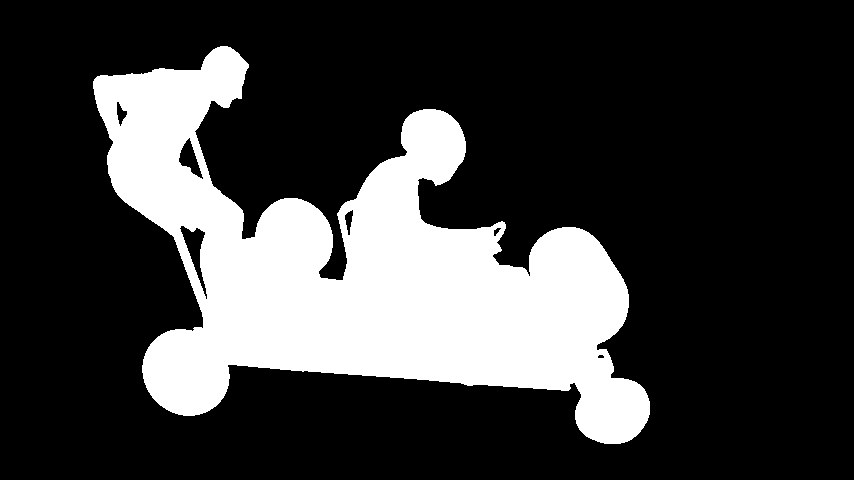} \\ 
\includegraphics[width=0.82in,height=0.5in]{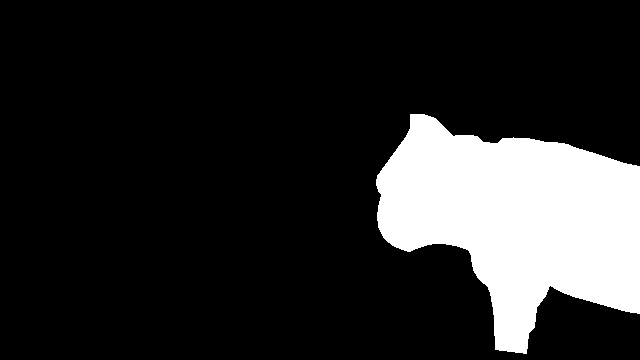}
\end{minipage}
}
\hspace{0.02cm}
\subfigure[Ours]{
\begin{minipage}[b]{0.10\textwidth}
\includegraphics[width=0.82in,height=0.5in]{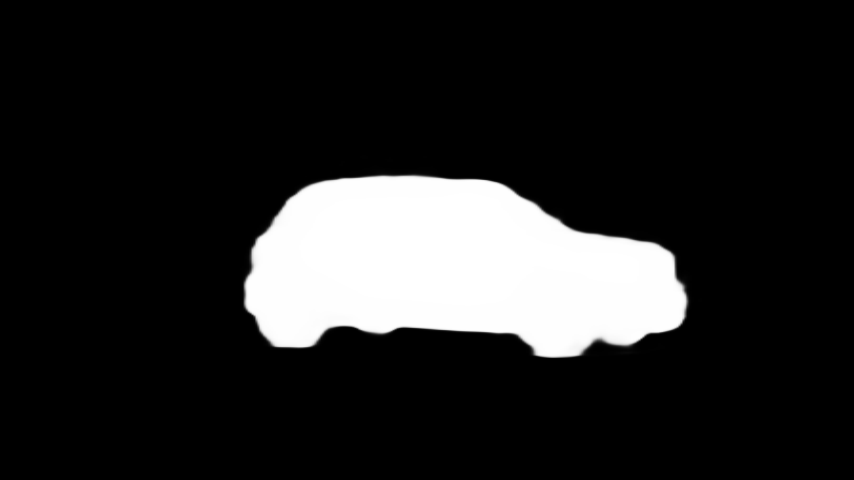} \\ 
\includegraphics[width=0.82in,height=0.5in]{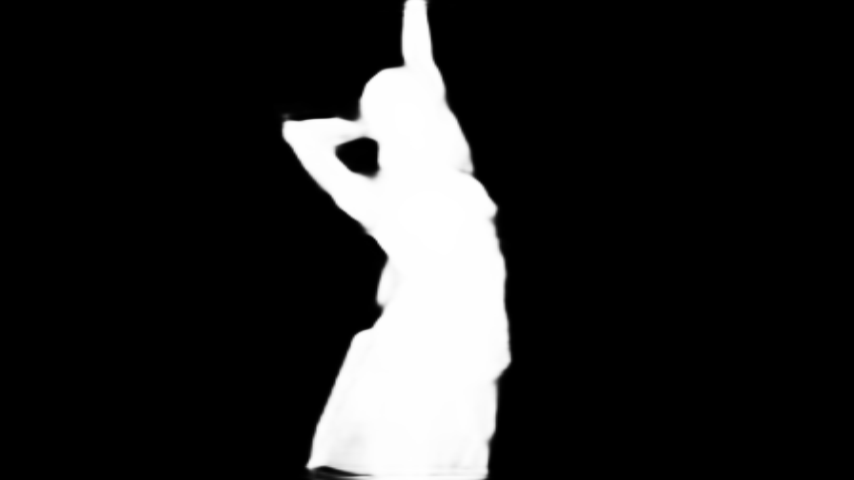} \\ 
\includegraphics[width=0.82in,height=0.5in]{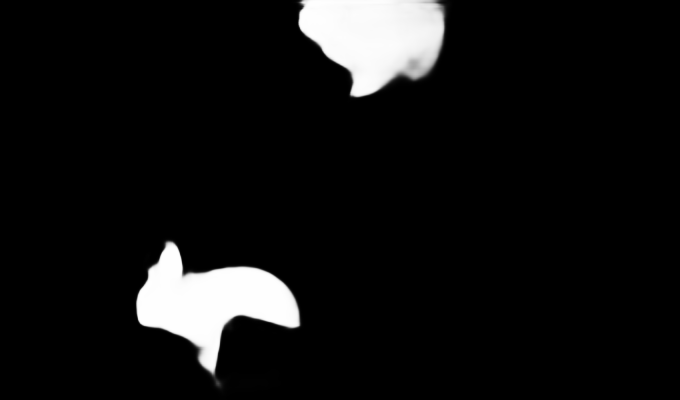} \\ 
\includegraphics[width=0.82in,height=0.5in]{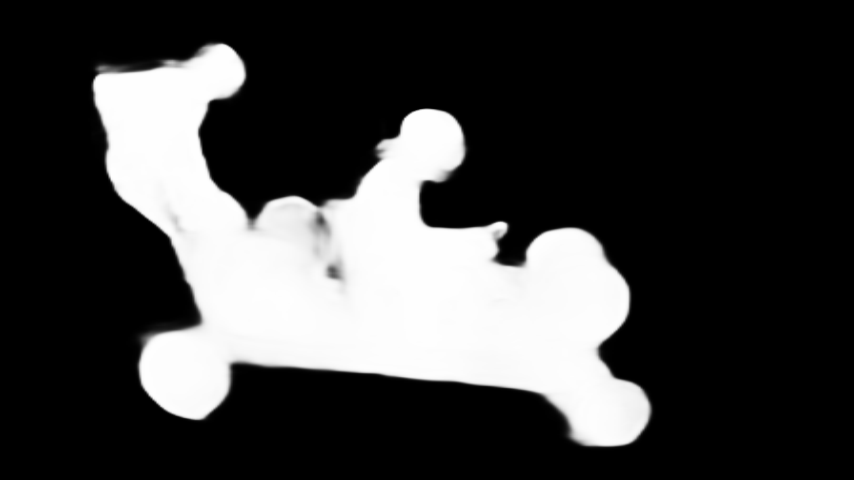} \\ 
\includegraphics[width=0.82in,height=0.5in]{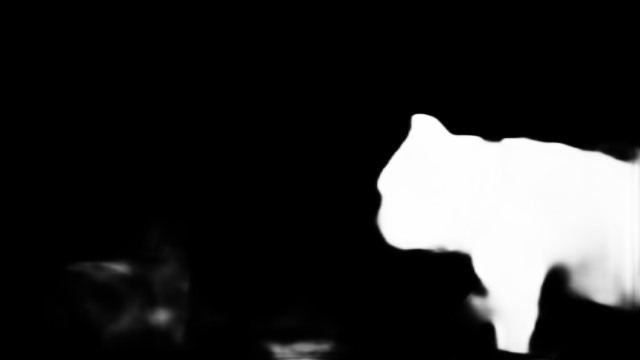}
\end{minipage}
}
\hspace{0.02cm}
\subfigure[PCSA]{
\begin{minipage}[b]{0.10\textwidth}
\includegraphics[width=0.82in,height=0.5in]{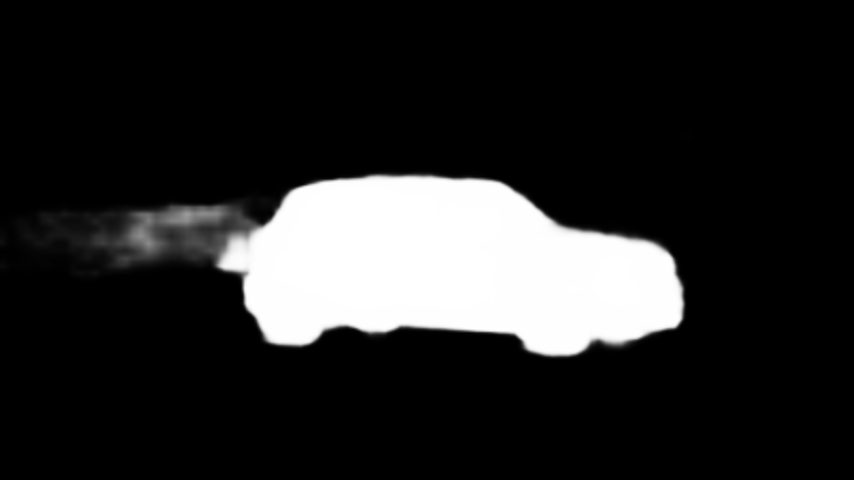} \\ 
\includegraphics[width=0.82in,height=0.5in]{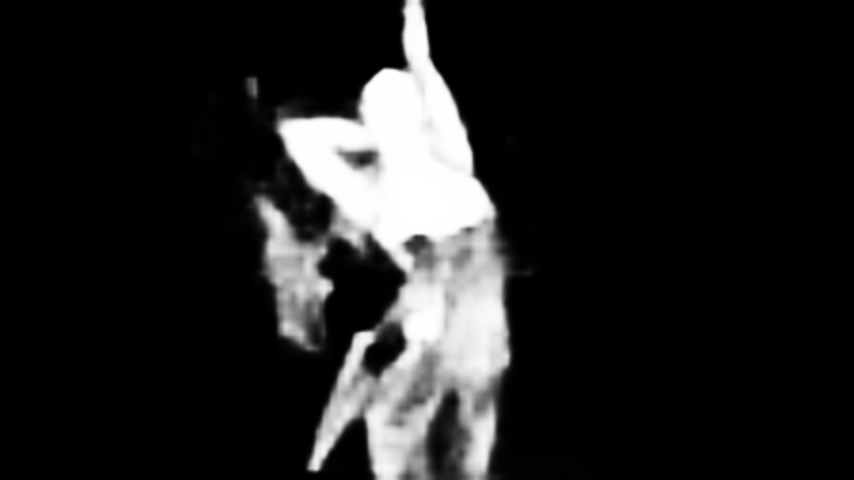} \\ 
\includegraphics[width=0.82in,height=0.5in]{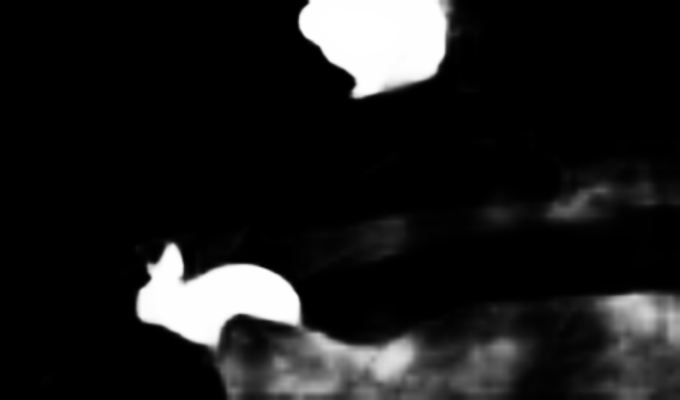} \\ 
\includegraphics[width=0.82in,height=0.5in]{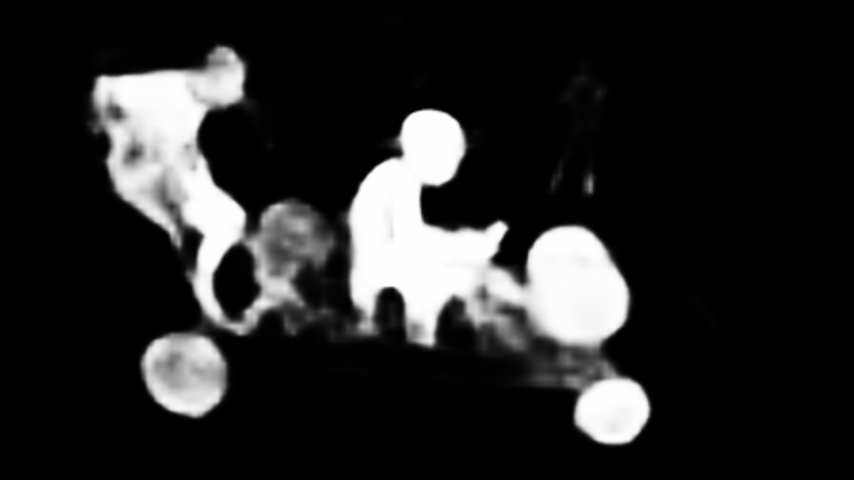} \\ 
\includegraphics[width=0.82in,height=0.5in]{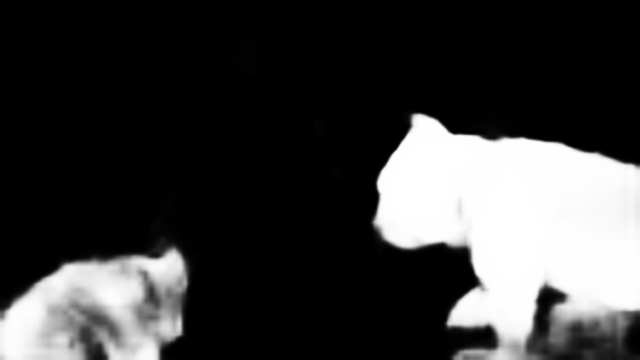}
\end{minipage}
}
\hspace{0.02cm}
\subfigure[MGA]{
\begin{minipage}[b]{0.10\textwidth}
\includegraphics[width=0.82in,height=0.5in]{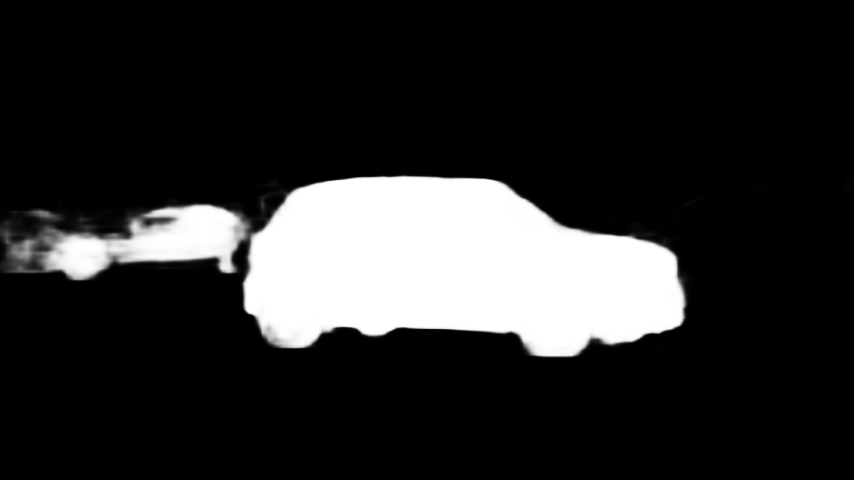} \\ 
\includegraphics[width=0.82in,height=0.5in]{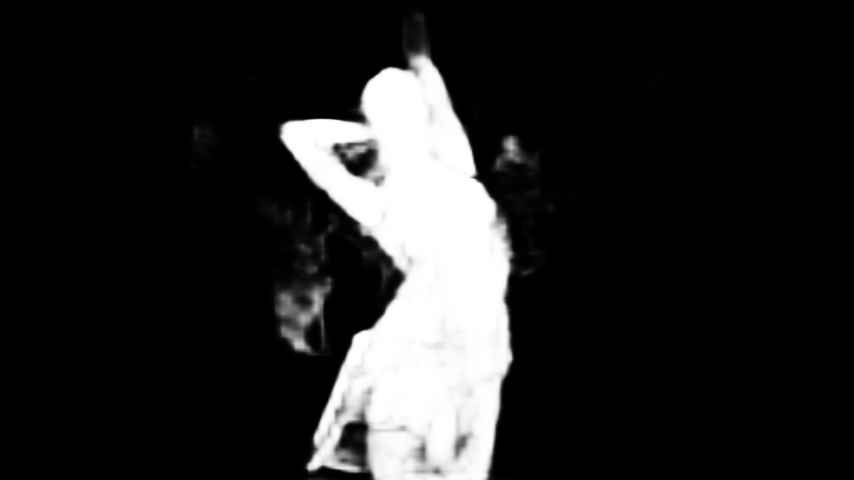} \\ 
\includegraphics[width=0.82in,height=0.5in]{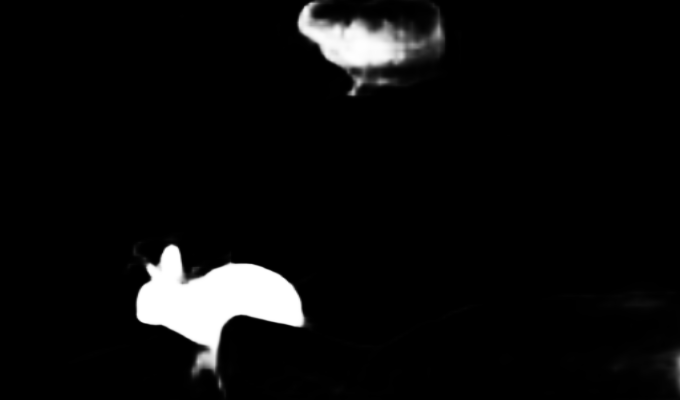} \\ 
\includegraphics[width=0.82in,height=0.5in]{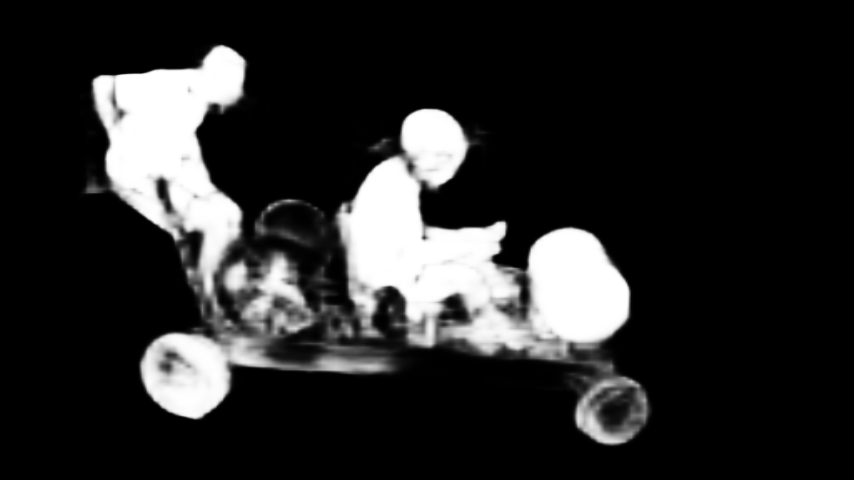} \\ 
\includegraphics[width=0.82in,height=0.5in]{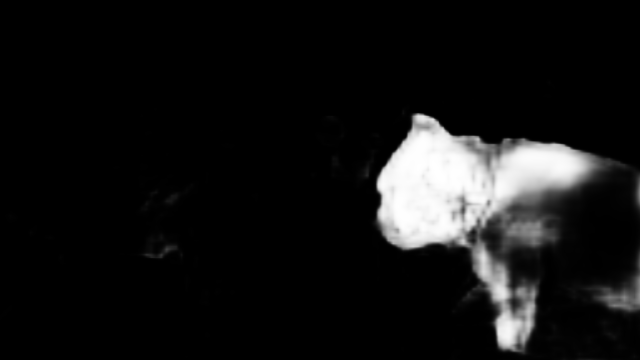}
\end{minipage}
}
\hspace{0.02cm}
\subfigure[SSAV]{
\begin{minipage}[b]{0.10\textwidth}
\includegraphics[width=0.82in,height=0.5in]{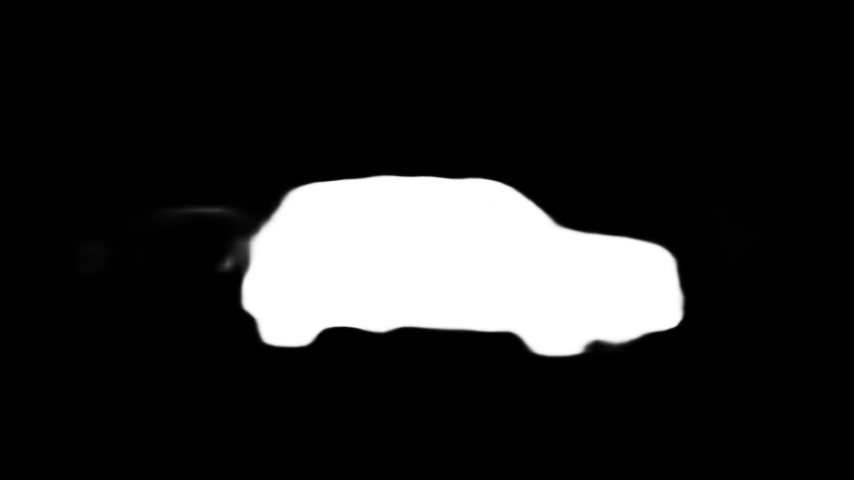} \\ 
\includegraphics[width=0.82in,height=0.5in]{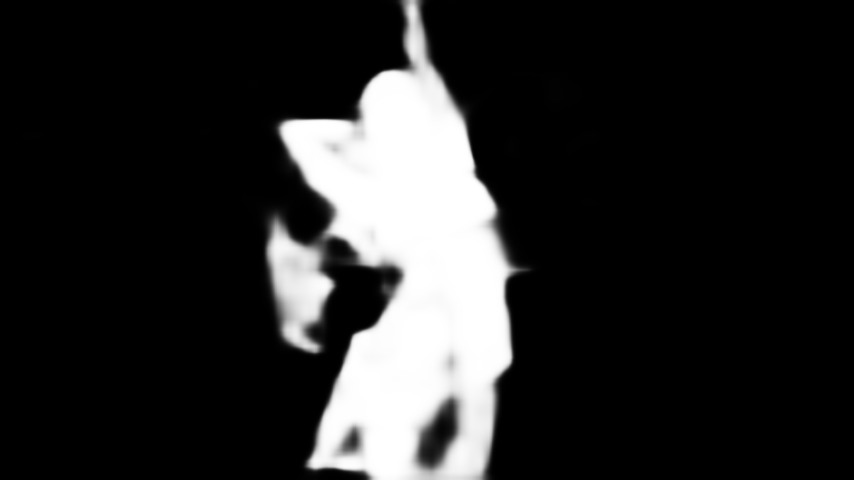} \\ 
\includegraphics[width=0.82in,height=0.5in]{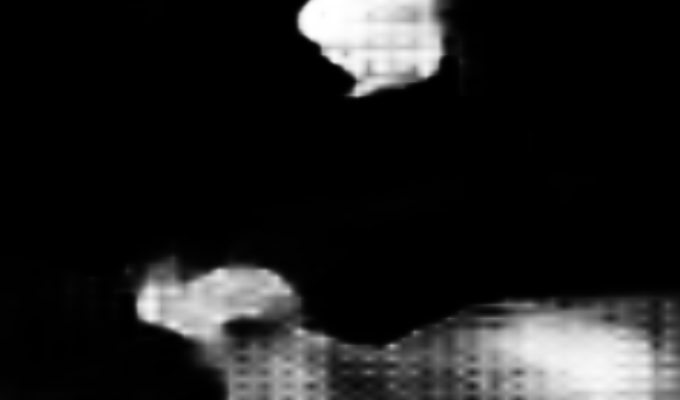} \\ 
\includegraphics[width=0.82in,height=0.5in]{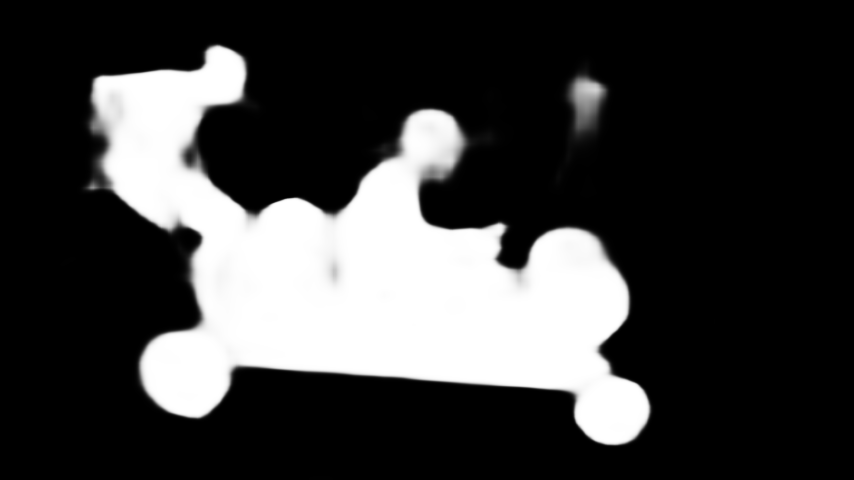} \\ 
\includegraphics[width=0.82in,height=0.5in]{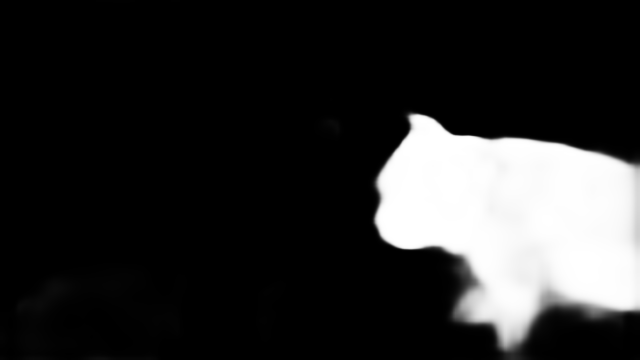}
\end{minipage}
}
\hspace{0.02cm}
\subfigure[FGRNE]{
\begin{minipage}[b]{0.10\textwidth}
\includegraphics[width=0.82in,height=0.5in]{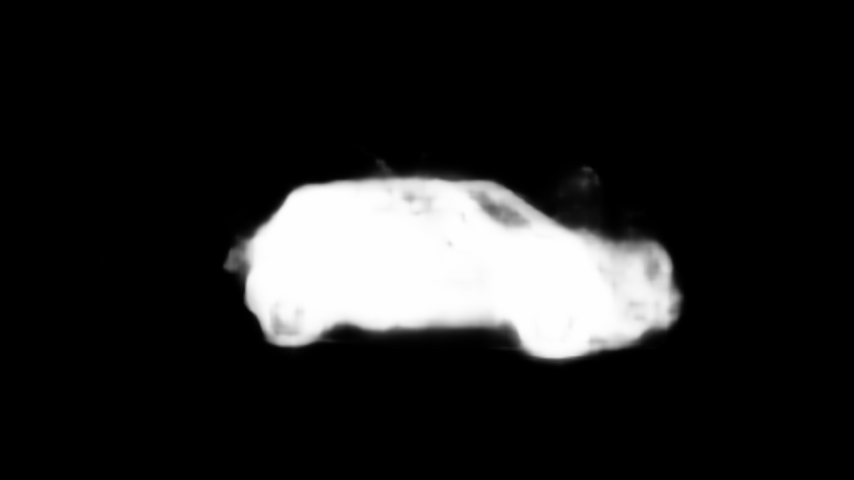} \\ 
\includegraphics[width=0.82in,height=0.5in]{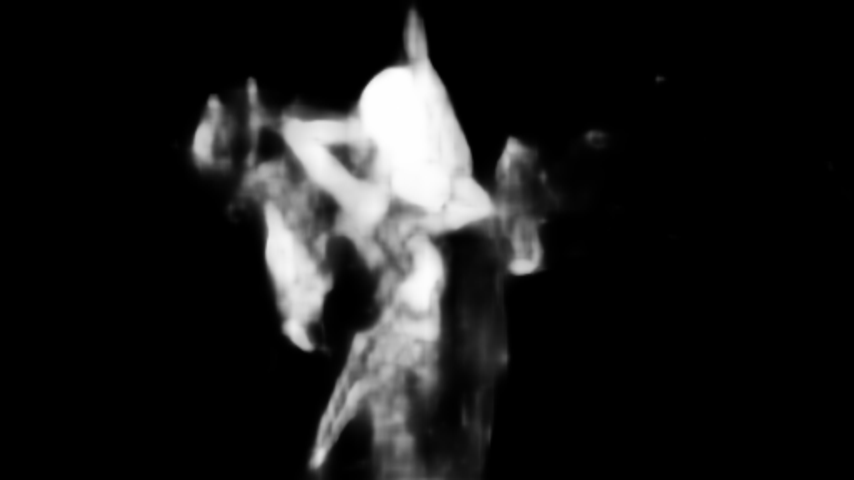} \\ 
\includegraphics[width=0.82in,height=0.5in]{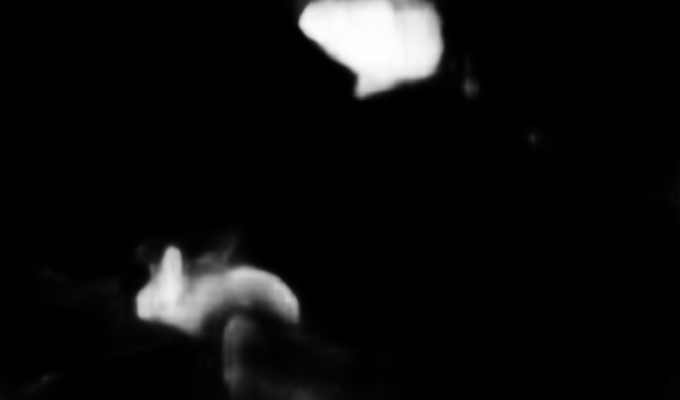} \\ 
\includegraphics[width=0.82in,height=0.5in]{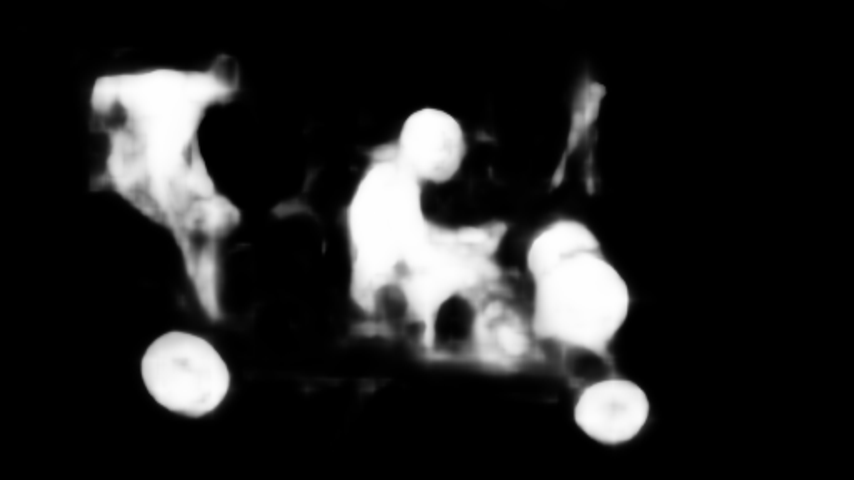} \\ 
\includegraphics[width=0.82in,height=0.5in]{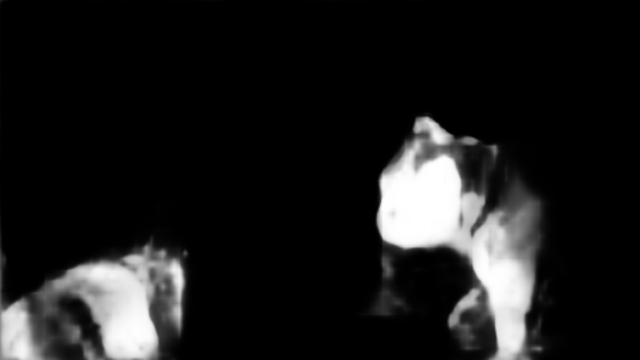}
\end{minipage}
}
\hspace{0.02cm}
\subfigure[ITSD]{
\begin{minipage}[b]{0.10\textwidth}
\includegraphics[width=0.82in,height=0.5in]{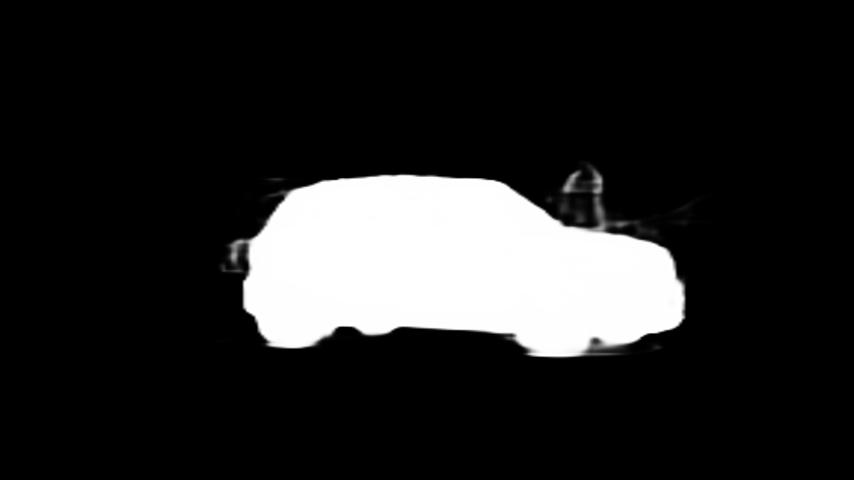} \\ 
\includegraphics[width=0.82in,height=0.5in]{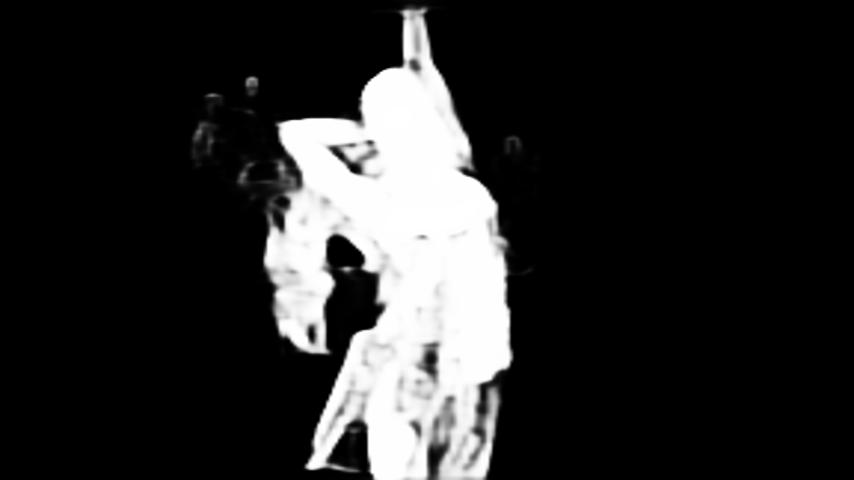} \\ 
\includegraphics[width=0.82in,height=0.5in]{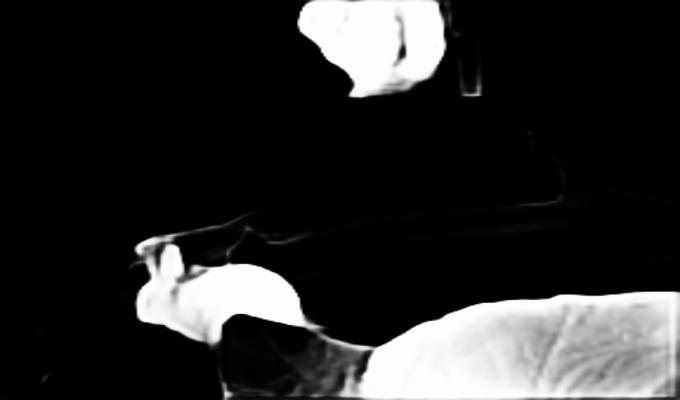} \\ 
\includegraphics[width=0.82in,height=0.5in]{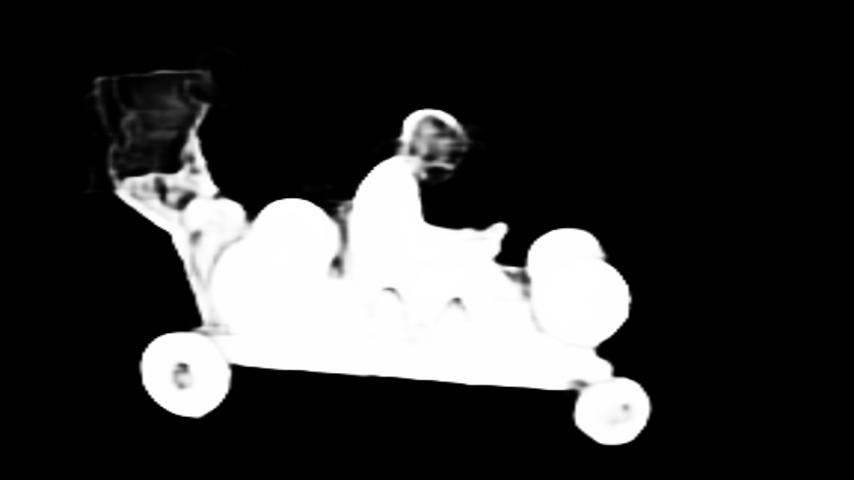} \\ 
\includegraphics[width=0.82in,height=0.5in]{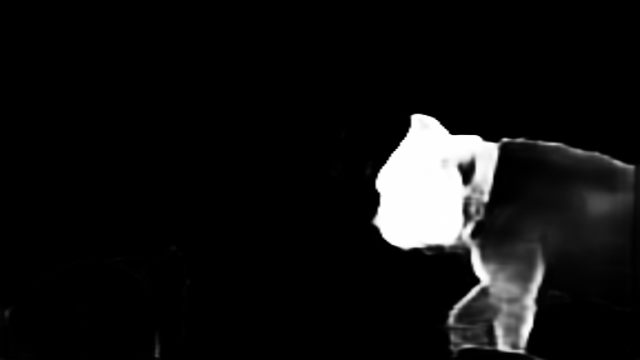}
\end{minipage}
}
\caption{Qualitative comparison of our method and the state-of-the-art methods.}
\label{fig:main}
\end{figure*}

\subsection{Loss Function}
% final predict
Following \cite{qin2019basnet}, an effective loss function is adopted, which is given by
% \begin{small}
\begin{eqnarray}
L_f = L_{BCE}(P_f, G) + L_{SSIM}(P_f, G) + L_{IoU}(P_f, G),
\end{eqnarray}
% \end{small}
where $L_{BCE}$, $L_{SSIM}$, $L_{IoU}$ are binary cross entropy loss, structural similarity index measure loss and intersection over union loss, respectively. $P_f$ and $G$ denote the predicted saliency map and ground truth, respectively. These three losses are given by
\begin{eqnarray}
&&\hspace{-2.6em}L_{BCE} = -(GlogP+(1-G)log(1-P)),
\\
&&\hspace{-3.0em}L_{SSIM} = 1 - \frac{(2\mu_P\mu_G+C_1)(2\sigma_{PG}+C_2)}{(\mu_P^2+\mu_G^2+C_1)(\sigma_P^2+\sigma_G^2+C_2)},
\\
&&\hspace{-2.1em}L_{IoU} = 1 - \frac{TP}{TN+TP+FP},
\end{eqnarray}
where $P$ denotes the predicted saliency map. $\mu_P$ and $\sigma_P$ ($\mu_G$ and $\sigma_G$) are the mean and standard deviations of the predicted saliency map (ground truth), respectively. $\sigma_{PG}$ is the covariance of $P$ and $G$. TP, TN and FP represent true-positive, true-negative and false-positive, respectively.

% for confidence estimation
For the CAG module at the i-th layer, we adopt the binary cross entropy loss to train the segmentation sub-network and $l_1$ loss to regress the confidence score prediction sub-network. Therefore, the loss function of CAG can be formulated as
% \begin{small}
\begin{eqnarray}
L_i^t = L_{BCE}(P_i^t, G_i) + L_{l_1}(s_i^t, \mathcal{F}_{IoU}(P_i^t, G_i)),
\end{eqnarray}
% \end{small}
where $\mathcal{F}_{IoU}$ is a function to calculate the IoU. $t$ indicates the type of features, including $RGB$ for RGB features and $OF$ for optical flow features. $G_i$ is downsampled from the ground truth, which has the consistent size as the predicted saliency map $P_i$.
% all
Finally, the total loss function can be written as
% \begin{small}
\begin{eqnarray}
L_{total} = L_f + \sum_{t}\sum_{i=1}^{N} L_i^t,
\end{eqnarray}
% \end{small}
where $N$ is 5.

\begin{table*}[t]
% \captionsetup{font={footnotesize},labelsep=period}
% \captionsetup{labelsep=period}
\caption{Quantitative comparison with the state-of-the-art static and video salient object detection methods by three evaluation metrics. The best three results are highlighted in \textcolor{red}{red}, \textcolor[rgb]{0,0.8,0}{green}, and \textcolor{blue}{blue} respectively.}
\label{tab:main}
% \footnotesize
\tabcolsep2pt %列宽
\renewcommand\arraystretch{1.1} %行距
\begin{center}
\begin{tabular}{c|c|ccc|ccc|ccc|ccc}
\hline
\multirow{2}{*}{Method} & \multirow{2}{*}{Year} & \multicolumn{3}{c|}{DAVSOD} & \multicolumn{3}{c|}{DAVIS} & \multicolumn{3}{c|}{SegV2} & \multicolumn{3}{c}{FBMS} \\ \cline{3-14}
 &  & max $F_\beta\uparrow$ & S $\uparrow$ & MAE $\downarrow$ & max $F_\beta\uparrow$ & S $\uparrow$ & MAE $\downarrow$ & max $F_\beta\uparrow$ & S $\uparrow$ & MAE $\downarrow$ & max $F_\beta\uparrow$ & S $\uparrow$ & MAE $\downarrow$ \\ \hline
\multicolumn{14}{c}{Static Salient Object Detection} \\ \hline
EGNet\cite{zhao2019egnet} & 2019 & 0.604 & 0.719 & 0.101 & 0.768 & 0.829 & 0.056 & 0.774 & 0.845 & \textcolor[rgb]{0,0.8,0}{0.024} & 0.848 & 0.878 & 0.044 \\
CPD\cite{wu2019cascaded} & 2019 & 0.613 & 0.723 & 0.092 & 0.810 & 0.863 & 0.031 & 0.778 & 0.841 & \textcolor{red}{0.023} & 0.841 & 0.867 & 0.048 \\
% PoolNet(CVPR2019) & 0.826 & 0.860 & 0.044 & 0.782 & 0.843 & 0.020 & 0.864 & 0.882 & 0.038 & 0.612 & 0.731 & 0.088 \\
BASNet\cite{qin2019basnet} & 2019 & 0.597 & 0.708 & 0.109 & 0.812 & 0.858 & 0.031 & 0.774 & 0.838 & \textcolor{red}{0.023} & 0.822 & 0.858 & 0.047 \\
ITSD\cite{zhou2020interactive} & 2020 & 0.651 & \textcolor{blue}{0.747} & 0.094 & 0.835 & 0.876 & 0.033 & \textcolor{blue}{0.807} & 0.787 & \textcolor{blue}{0.027} & 0.843 & 0.869 & \textcolor{blue}{0.040} \\ \hline
\multicolumn{14}{c}{Video Salient Object Detection} \\ \hline
STBP\cite{xi2016salient} & 2016 & 0.410 & 0.568 & 0.160 & 0.544 & 0.677 & 0.096 & 0.640 & 0.735 & 0.061 & 0.595 & 0.627 & 0.152 \\
SFLR\cite{chen2017video} & 2017 & 0.478 & 0.624 & 0.132 & 0.727 & 0.790 & 0.056 & 0.745 & 0.804 & 0.037 & 0.660 & 0.699 & 0.117 \\
SCOM\cite{chen2018scom} & 2018 & 0.464 & 0.599 & 0.220 & 0.783 & 0.832 & 0.048 & 0.764 & 0.815 & 0.030 & 0.797 & 0.794 & 0.079 \\
SCNN\cite{tang2018weakly} & 2018 & 0.532 & 0.674 & 0.128 & 0.714 & 0.783 & 0.064 & - & - & - & 0.762 & 0.794 & 0.095 \\
FGRNE\cite{li2018flow} & 2018 & 0.573 & 0.693 & 0.098 & 0.783 & 0.838 & 0.043 & - & - & - & 0.767 & 0.809 & 0.088 \\
PDBM\cite{song2018pyramid} & 2018 & 0.573 & 0.698 & 0.116 & 0.855 & 0.882 & \textcolor{blue}{0.028} & 0.800 & \textcolor{blue}{0.864} & \textcolor[rgb]{0,0.8,0}{0.024} & 0.821 & 0.851 & 0.064 \\
SSAV\cite{fan2019shifting} & 2019 & 0.603 & 0.724 & 0.092 & 0.861 & 0.893 & \textcolor{blue}{0.028} & 0.801 & 0.851 & \textcolor{red}{0.023} & \textcolor[rgb]{0,0.8,0}{0.865} & \textcolor[rgb]{0,0.8,0}{0.879} & \textcolor{blue}{0.040} \\
MGA\cite{li2019motion} & 2019 & \textcolor{blue}{0.655} & \textcolor[rgb]{0,0.8,0}{0.751} & \textcolor[rgb]{0,0.8,0}{0.081} & \textcolor[rgb]{0,0.8,0}{0.892} & \textcolor{red}{0.912} & \textcolor[rgb]{0,0.8,0}{0.022} & - & - & - & \textcolor{red}{0.906} & \textcolor{red}{0.910} & \textcolor{red}{0.026} \\
PCSA\cite{2020Pyramid} & 2020 & \textcolor[rgb]{0,0.8,0}{0.656} & 0.741 & \textcolor{blue}{0.086} & \textcolor{blue}{0.878} & \textcolor{blue}{0.901} & \textcolor[rgb]{0,0.8,0}{0.022} & \textcolor[rgb]{0,0.8,0}{0.811} & \textcolor{red}{0.866} & \textcolor[rgb]{0,0.8,0}{0.024} & 0.831 & 0.864 & \textcolor{blue}{0.040} \\
% TENet(ECCV2020) & 0.880 & 0.904 & 0.017 & 0.809 & 0.866 & 0.026 & 0.904 & 0.910 & 0.025 & 0.695 & 0.779 & 0.070 \\
Ours & - & \textcolor{red}{0.670} & \textcolor{red}{0.762} & \textcolor{red}{0.072} & \textcolor{red}{0.898} & \textcolor[rgb]{0,0.8,0}{0.906} & \textcolor{red}{0.018} & \textcolor{red}{0.826} & \textcolor[rgb]{0,0.8,0}{0.865} & \textcolor{blue}{0.027} & \textcolor{blue}{0.858} & \textcolor{blue}{0.870} & \textcolor[rgb]{0,0.8,0}{0.039} \\
\hline
\end{tabular}
\end{center}
\end{table*}

\begin{table}[t]
\begin{center}
\caption{Ablation studies of the proposed network architecture on DAVIS and DAVSOD datasets.}
\label{tab:ablation}
\small
\begin{tabular}{c|c|c|c|c}
  \hline
  \multirow{2}{*}{Methods} & \multicolumn{2}{c|}{DAVIS} & \multicolumn{2}{c}{DAVSOD} \\ \cline{2-5}
  & max $F_\beta\uparrow$ & MAE $\downarrow$ & max $F_\beta\uparrow$ & MAE $\downarrow$ \\ \hline
  Baseline (Concat) & 0.869 & 0.023 & 0.645 & 0.080 \\ \hline
%   Add & 0.865 & 0.024 & 0.656 & 0.082 \\ \hline
%   Mul & 0.870 & 0.023 & 0.625 & 0.087 \\ \hline
  Baseline + CAG & 0.887 & 0.022 & 0.662 & 0.072 \\ \hline
  Baseline + DDE & 0.892 & 0.020 & 0.656 & 0.074 \\ \hline
  Ours & 0.898 & 0.018 & 0.670 & 0.072 \\ \hline
\end{tabular}
\end{center}
\end{table}

\section{Experiments}
\subsection{Experimental Setup}
\textbf{Datasets:} To evaluate the effectiveness of our method, we conduct experiments on four widely used public datasets, including SegV2 \cite{li2013video}, FBMS \cite{ochs2013segmentation}, DAVIS \cite{perazzi2016benchmark} and DAVSOD \cite{fan2019shifting} datasets. 
% SegV2 contains 13 video sequences with various instances including animals, vehicles and humans. FBMS contains 59 video sequences including 720 annotated frames. DAVIS contains 50 video sequences with 3455 high quality pixel-wise annotated frames, most of which involve diverse content with occlusion and ambiguous boundary. DAVSOD is the most challenging dataset, which contains 226 video sequences with 23938 frames with diverse realistic-scenes, instances and motion patterns.

\textbf{Evaluation Metrics:} To quantitatively evaluate the performance of VSOD, we adopt three metrics in our experiments, i.e., F-measure \cite{achanta2009frequency}, S-measure \cite{fan2017structure} and Mean Absolute Error (MAE) \cite{perazzi2012saliency}. F-measure is defined as
% \begin{small}
\begin{eqnarray}
F_\beta = \frac{(1+\beta^2) \times Precision \times Recall}{\beta^2 \times Precision + Recall},
\end{eqnarray}
% \end{small}
where $\beta^2$ is set to 0.3 and we report the maximum F-measure for evaluation. S-measure takes both region-aware and object-aware structural similarity into consideration. MAE measures the pixel-level average absolute difference between the predicted map and ground truth.
% , which can be computed by
% % \begin{small}
% \begin{eqnarray}
% S = \mu * S_o + (1 - \mu) * S_r,
% \end{eqnarray}
% % \end{small}
% where $\mu$ is set to 0.5. $S_o$ and $S_r$ denote object-aware and region-aware structural similarity measures. MAE measures the pixel-level average absolute difference between the predicted saliency map $P$ and the ground truth $G$:
% \begin{eqnarray}
% MAE = \frac{1}{n}\sum_{i=1}^{n}|P_i - G_i|,
% \end{eqnarray}
% where $n$ indicates the total number of pixels.

\textbf{Implementation Details:} We implement our method on Pytorch. We use the pre-trained ResNet-101 \cite{he2016deep} as our initial backbone. We employ RAFT \cite{teed2020raft} to render optical flow images. Following the previous methods \cite{li2019motion,2020Pyramid}, we remove the CAG modules and the DDE modules, and pre-train our model with the training set of DUTS \cite{wang2017learning}. After pre-training, we use the training set of DAVIS and DAVSOD to train the whole network. We resize the input images to 448 $\times$ 448. We apply random horizontal flipping and scaling the input images with scales \{0.75, 1, 1.25\}. We use the Adam optimizer to train our model with an initial learning rate of 1e-5 with batch size of 4.

\subsection{Comparison with State-of-the-art Methods}
We compare our method against four state-of-the-art static salient object detection methods including EGNet \cite{zhao2019egnet}, CPD \cite{wu2019cascaded}, BASNet \cite{qin2019basnet}, ITSD \cite{zhou2020interactive}. We also compare our method against nine state-of-the-art video salient object detection methods including STBP \cite{xi2016salient}, SFLR \cite{chen2017video}, SCOM \cite{chen2018scom}, SCNN \cite{tang2018weakly}, FGRNE \cite{li2018flow}, PDBM \cite{song2018pyramid}, SSAV \cite{fan2019shifting}, MGA \cite{li2019motion}, PCSA \cite{2020Pyramid}. In our experiments, we employ the evaluation code \cite{fan2019shifting} to evaluate all the saliency maps for a fair comparison.

\textbf{Quantitative Evaluation.} We first conduct a quantitative evaluation, as shown in Table \ref{tab:main}. It is observed that our method achieves state-of-the-art results on three datasets. Specifically, our method outperforms other state-of-the-art methods under the metric max $F_\beta$ on DAVIS, SegV2 and DAVSOD datasets, and achieves the third-best performance on FBMS dataset. Note that the leading performance of MGA on FBMS dataset is mainly induced by using the FBMS training set. Notably, our method achieves significant performance improvement on the most challenging dataset DAVSOD compared with the second-best method MGA (0.670 V.S. 0.655, 0.762 V.S. 0.751, 0.072 V.S. 0.081 in terms of max $F_\beta$, S and MAE respectively), which demonstrates the superior performance of our method in complex scenes.

\textbf{Qualitative Evaluation.} We conduct a qualitative evaluation in different scenes, as shown in Figure \ref{fig:main}. The results show that our method can accurately locate and segment the salient objects in several complex scenes, such as multiple moving objects (the first row), cluttered foreground and background (the second and third rows), complex boundary (the fourth row) and saliency shifts (the last row).

\subsection{Ablation Studies}
The proposed network is composed of two main modules, the Confidence-guided Adaptive Gate (CAG) module and the Dual Differential Enhancement (DDE) module. To verify the effectiveness of each component, we conduct an ablation experiment on two large-scale datasets, i.e., DAVIS and DAVSOD. The experimental results are shown in Table \ref{tab:ablation}. The first row is the baseline model that merges two features with the concatenation operation. The second and third rows show that both CAG and DDE can boost performance compared with the baseline. Moreover, the combination of the CAG and DDE modules can further improve the performance.
% Specifically, the baseline model with the CAG performs better with DDE on the DAVSOD dataset, which covers complex motion patterns and various images of different resolutions and illuminations. It demonstrates the superiority of the CAG to suppress noisy information. On the other hand, due to the images of high quality and stable motion from the DAVIS dataset, it is important to make full use of both spatial and temporal information. As a result, the baseline with DDE outperforms that with CAG on the DAVIS dataset.

% Furthermore, we compare our method with some naive fusion techniques including concatenation, element-wise addition and element-wise multiplication, denoted as Concat, Add and Mul respectively. Experimental results including our full method are shown in Table \ref{tab:fusion}. 

\begin{table}[t]
\begin{center}
\caption{Comparison with different fusion mechanisms on the DAVIS and DAVSOD datasets.}
\label{tab:fusion}
\small
\begin{tabular}{c|c|c|c|c}
  \hline
  \multirow{2}{*}{Methods} & \multicolumn{2}{c|}{DAVIS} & \multicolumn{2}{c}{DAVSOD} \\ \cline{2-5}
  & max $F_\beta\uparrow$ & MAE $\downarrow$ & max $F_\beta\uparrow$ & MAE $\downarrow$ \\ \hline
  Cat & 0.869 & 0.023 & 0.645 & 0.080 \\ \hline
  Add & 0.865 & 0.024 & 0.656 & 0.082 \\ \hline
  Mul & 0.870 & 0.023 & 0.625 & 0.087 \\ \hline
  Ours & 0.898 & 0.018 & 0.670 & 0.072 \\ \hline
\end{tabular}
\end{center}
\end{table}

\section{Conclusions}
% In this paper, we propose a confidence-guided gated network for video salient object detection in this paper, which explore the distinctive information from the spatial and temporal branches adaptively. We use the adaptive gate modules to re-calibrated the noisy information according to the confidence predicted by the confidence score estimator. In addition, we propose the feature interaction modules to learn comprehensive information from two complementary branches. Experimental results demonstrate that our method achieves state-of-the-art results on four widely used datasets.

%for  network to address two challenges in video salient object detection, i.e., the robustness to unreliable information and the completeness of the salient object

In this paper, we propose a new video salient object detection method to exploit available spatial and temporal information to predict robust saliency maps. We first propose a Confidence-guided Adaptive Gate (CAG) module to filters unreliable cues from spatial and temporal information. We then propose a Dual Differential Enhancement (DDE) module, which merges spatial and temporal information enhanced by differential features to capture complementary information. Experimental results on four widely used datasets demonstrate the effectiveness of our proposed method.

\bibliographystyle{IEEEbib}
\bibliography{icme2021template}

\end{document}